\documentclass{article}

\usepackage{iclr2027_conference,times}

\usepackage[utf8]{inputenc}
\usepackage[T1]{fontenc}
\usepackage{microtype}

\usepackage{amsmath}
\usepackage{amssymb}
\usepackage{amsthm}
\usepackage{mathtools}
\usepackage{bm}
\usepackage{nicefrac}

\usepackage{graphicx}
\usepackage{booktabs}
\usepackage{multirow}
\usepackage{makecell}
\usepackage{tabularx}
\usepackage{array}
\usepackage{subcaption}
\usepackage{wrapfig}
\usepackage{algorithm}
\usepackage{algorithmic}
\usepackage{placeins}

\usepackage{xspace}
\usepackage{xcolor}
\usepackage{soul}
\usepackage{enumitem}
\usepackage{url}
\usepackage{hyperref}
\usepackage[nameinlink,capitalise]{cleveref}

\definecolor{comemblue}{RGB}{45,95,155}
\definecolor{comemorange}{RGB}{214,112,45}
\definecolor{comemgray}{RGB}{242,242,242}

\setlist[itemize]{leftmargin=1.35em,labelsep=0.45em,topsep=0.25em,itemsep=0.15em}
\setlist[enumerate]{leftmargin=1.55em,labelsep=0.45em,topsep=0.25em,itemsep=0.15em}

\usepackage{amsmath,amsfonts,bm}

\def\eqref#1{equation~\ref{#1}}

\def\1{\bm{1}}

\DeclareMathAlphabet{\mathsfit}{\encodingdefault}{\sfdefault}{m}{sl}
\SetMathAlphabet{\mathsfit}{bold}{\encodingdefault}{\sfdefault}{bx}{n}

\newcommand{\comembench}{\textsc{CoMemBench}\xspace}

\newcommand{\VNCR}{\textsc{VNCR}\xspace}
\newcommand{\VHS}{\textsc{VHS}\xspace}
\newcommand{\ICS}{\textsc{ICS}\xspace}

\title{\comembench: Benchmarking Collaborative Memory Boundaries across Multi-Agent Workflow Topologies}

\renewcommand{\thefootnote}{\fnsymbol{footnote}}
\author{
\textbf{Sen Zhao\textsuperscript{1}\thanks{Equal contribution.} \quad
Ruiqi Kong\textsuperscript{1}\footnotemark[1] \quad
Zuyu Zhang\textsuperscript{1} \quad
Lifeng Shen\textsuperscript{2}} \\
\textbf{Xinyu He\textsuperscript{4} \quad
Ding Zou\textsuperscript{5} \quad
Xu Zhang\textsuperscript{1}\thanks{Corresponding author: \texttt{zhangx@cqupt.edu.cn}.} \quad
Qinghua Zhang\textsuperscript{3}} \\[0.35em]
{\normalfont\small \textsuperscript{1}Academy of Advanced Interdisciplinary Studies,} \\
{\normalfont\small Chongqing University of Posts and Telecommunications, Chongqing, China} \\
{\normalfont\small \textsuperscript{2}School of Artificial Intelligence,} \\
{\normalfont\small Chongqing University of Posts and Telecommunications, Chongqing, China} \\
{\normalfont\small \textsuperscript{3}School of Computer Science and Technology,} \\
{\normalfont\small Chongqing University of Posts and Telecommunications, Chongqing, China} \\
{\normalfont\small \textsuperscript{4}Towngas, China} \\
{\normalfont\small \textsuperscript{5}Intelligent System Department, Zhongxing Telecom Equipment (ZTE),} \\
{\normalfont\small Changsha, Hunan, China}
}

\iclrfinalcopy

\begin{document}

\maketitle
\renewcommand{\thefootnote}{\arabic{footnote}}
\setcounter{footnote}{0}
\lhead{Under review}

\begin{abstract}
Multi-agent workflows require task-relevant information to be shared across agents, while irrelevant, stale, unverified, or incompatible information must remain isolated. We call this task-conditioned scope of information a \emph{collaborative memory boundary}. Workflow topology determines which intermediate artifacts are applicable to which downstream workers and when they cease to be valid, thereby providing a structural stress dimension for sharing and isolation. Existing memory benchmarks primarily evaluate retention and retrieval, whereas multi-agent benchmarks emphasize coordination and end-to-end completion, leaving topology-conditioned memory boundaries largely unmeasured. We introduce \comembench, an execution-grounded benchmark for collaborative memory sharing and isolation across multi-agent workflow topologies. It constructs 800 composite workflows across four domains from source-grounded dependency graphs, with node-local specifications, verifiable artifact handoffs, native evaluators, and matched isolation challenges. \comembench measures workflow completion, verified node progress, required-handoff reliability, isolation robustness, and token cost. Experiments reveal a sharing--isolation trade-off: broader context improves information availability but can weaken isolation, while system rankings shift across topologies and artifact violations. \comembench is released at http://comembench.world.

\end{abstract}

\section{Introduction}
\label{sec:introduction}

{\color{black}Large language model agents increasingly tackle complex tasks through teams of specialized workers~\citep{zhu2025multiagentbench,ke2026masorchestra}. These workers act on different instructions, observations, and execution histories, making selective information reuse central to collaboration. A downstream task must receive the evidence, state, or intermediate result it requires, while contextually inapplicable information---such as a sibling-branch result, an unverified output, or a superseded state---must not alter its execution. We call this task- and state-dependent distinction a \emph{collaborative memory boundary}. Maintaining such a boundary has two requirements: required information must remain available to the appropriate consumer, and inapplicable information must not change that consumer's verified behavior.\par}

\begin{wrapfigure}[22]{r}{0.65\textwidth}
    \vspace{-1.35em}
    \centering
    \includegraphics[width=\linewidth]{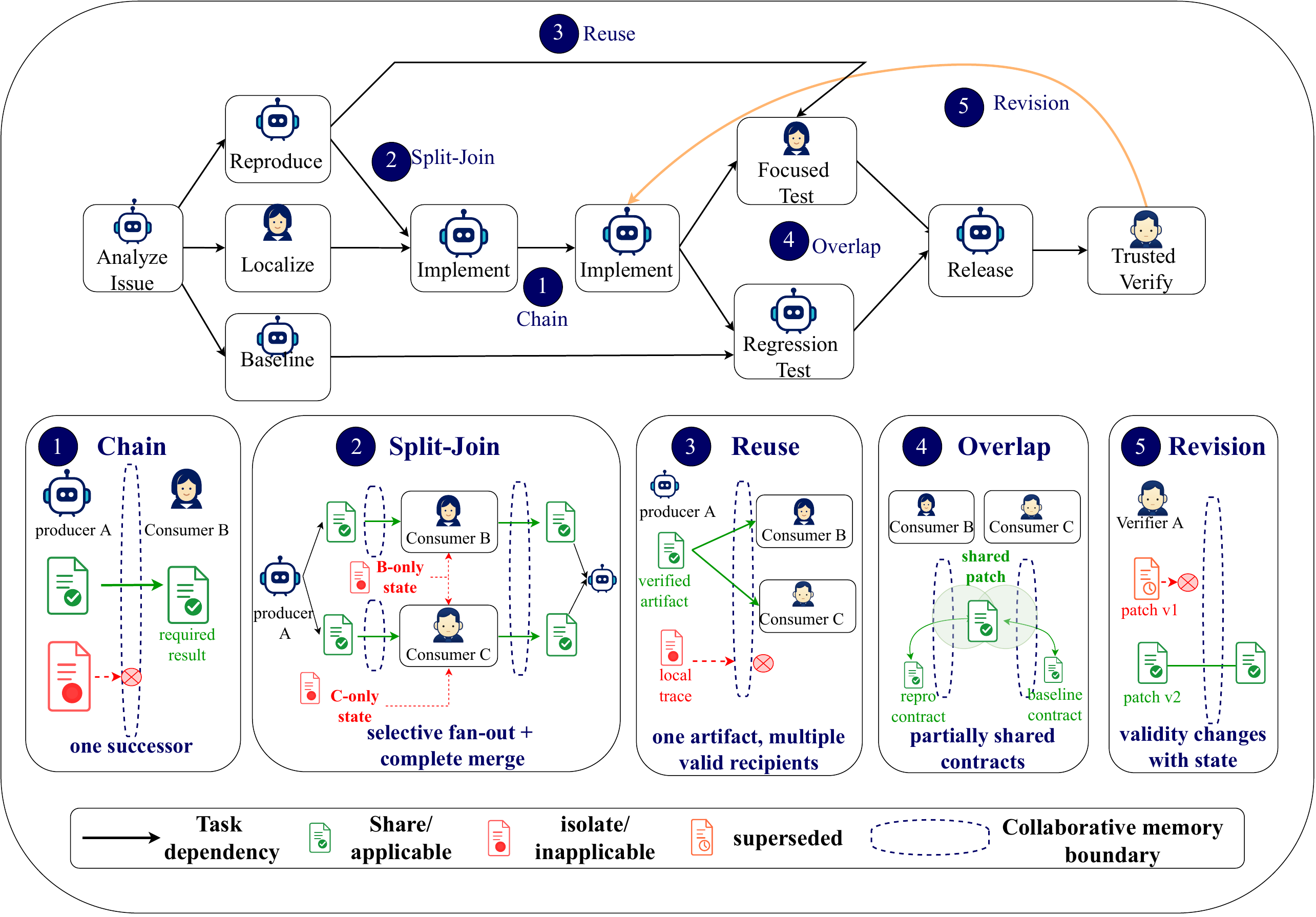}
    \caption{\textcolor{black}{\textbf{Topology-conditioned memory boundaries.} Five dependency motifs change where workflow records are required, branch-specific, reused, partially shared, or superseded.}}
    \label{fig:memory-boundary-motivation}
    \vspace{-1.2em}
\end{wrapfigure}

{\color{black}Workflow topology makes this boundary vary across consumers and execution states. Figure~\ref{fig:memory-boundary-motivation} illustrates five recurring cases. A chain transfers a required result to its successor; a split--join keeps branch-specific records separate before combining their outputs; reuse makes one verified result relevant to several consumers; overlap creates both shared and branch-specific dependencies; and bounded revision supersedes an earlier result after verification failure. These motifs do not define different kinds of memory; they change the consumers, workflow states, and dependency regions under which the same record is applicable. The same intermediate result may therefore be required along one edge, inapplicable to a sibling branch, reusable by several consumers, or superseded after revision. Our central question is whether multi-agent systems can maintain both sides of the collaborative memory boundary as these structural conditions change.\par}

{\color{black}Existing benchmarks cover complementary aspects of this problem, including long-context memory, governed or privacy-constrained group memory, multi-agent orchestration, and executable tool use~\citep{maharana2024locomo,wu2025longmemeval,hu2026memoryagentbench,yang2026groupmembench,ren2026gatemem,park2026pacbench,zhu2025multiagentbench,shen2024taskbench,ke2026masorchestra,li2025toolathlon}. As Table~\ref{tab:benchmark-comparison} shows, they do not jointly evaluate whether systems deliver required information and resist contextually inapplicable information across topology-diverse, executable workflows. A checkmark denotes a capability explicitly instantiated and evaluated. End-to-end success alone cannot distinguish a missing required handoff from an otherwise valid consumer being altered by an inappropriate record.\par}

{\color{black}We introduce \comembench, an execution-grounded benchmark of 800 source-grounded workflows across four domains. \comembench makes three contributions: \textbf{benchmark formulation}, operationalizing collaborative memory boundaries through declared handoffs and matched isolation interventions across diverse workflow topologies; \textbf{dataset and construction}, providing executable workflows with node-local specifications, verifiable artifact handoffs, and native evaluators; and \textbf{evaluation and findings}, separating terminal completion, verified progress, required sharing, and isolation robustness across systems, backbones, and memory mechanisms. Experiments show that broader information availability can improve progress without preserving isolation, while system behavior varies across workflow topology and artifact-violation type.\par}

\section{Related Work}
\label{sec:related_work}

\subsection{Agent Memory and Information Boundaries}

{\color{black}Memory evaluation has expanded from fixed-corpus evidence use~\citep{bai2023longbench,hsieh2024ruler,petroni2021kilt,thakur2021beir} to persistent and interactive settings. MemoryBank, LoCoMo, LongMemEval, RealTalk, and StoryBench evaluate long-term or multi-session recall and reasoning~\citep{zhong2023memorybank,maharana2024locomo,wu2025longmemeval,lee2025realtalk,wan2025storybench}, while MemoryAgentBench adds incremental ingestion, test-time learning, and selective forgetting~\citep{hu2026memoryagentbench}. These benchmarks move beyond static reading but primarily study memory accumulated and used by one focal agent, rather than its selective availability among agents executing interdependent tasks.\par}

\clearpage
\begin{table}[!t]
    \centering
    \small
    \setlength{\tabcolsep}{4.2pt}
    \renewcommand{\arraystretch}{1.12}
    \caption{Comparison with related benchmarks.}
    \label{tab:benchmark-comparison}
    \resizebox{\textwidth}{!}{%
    \begin{tabular}{@{}lcccccc@{}}
        \toprule
        \textbf{Benchmark} &
        \makecell{\textbf{Multi-Agent}\\\textbf{Workflow}} &
        \makecell{\textbf{Workflow Topology}\\\textbf{Diversity}} &
        \makecell{\textbf{Artifact}\\\textbf{Handoffs}} &
        \makecell{\textbf{Sharing}\\\textbf{Evaluation}} &
        \makecell{\textbf{Isolation}\\\textbf{Evaluation}} &
        \makecell{\textbf{Containerized}\\\textbf{Execution}} \\
        \midrule
        LoCoMo~\citep{maharana2024locomo}                    & -- & -- & -- & -- & -- & -- \\
        LongMemEval~\citep{wu2025longmemeval}                & -- & -- & -- & -- & -- & -- \\
        MemoryAgentBench~\citep{hu2026memoryagentbench}      & -- & -- & -- & -- & -- & -- \\
        GroupMemBench~\citep{yang2026groupmembench}          & -- & -- & -- & -- & -- & -- \\
        GateMem~\citep{ren2026gatemem}                       & -- & -- & -- & -- & $\checkmark$ & -- \\
        PAC-Bench~\citep{park2026pacbench}                   & $\checkmark$ & -- & -- & -- & $\checkmark$ & -- \\
        PeopleJoin~\citep{jhamtani2025peoplejoin}            & -- & -- & -- & -- & -- & -- \\
        SILO-BENCH~\citep{zhang2026silobench}                & $\checkmark$ & $\checkmark$ & $\checkmark$ & -- & -- & -- \\
        MultiAgentBench~\citep{zhu2025multiagentbench}       & $\checkmark$ & -- & -- & -- & -- & -- \\
        TaskBench~\citep{shen2024taskbench}                  & -- & $\checkmark$ & -- & -- & -- & -- \\
        MASBench~\citep{ke2026masorchestra}                  & $\checkmark$ & $\checkmark$ & $\checkmark$ & -- & -- & -- \\
        Toolathlon~\citep{li2025toolathlon}                  & -- & -- & -- & -- & -- & $\checkmark$ \\
        \midrule
        \textbf{\comembench}                                 & $\checkmark$ & $\checkmark$ & $\checkmark$ & $\checkmark$ & $\checkmark$ & $\checkmark$ \\
        \bottomrule
    \end{tabular}%
    }
\end{table}

{\color{black}GroupMemBench evaluates speaker-grounded memory~\citep{yang2026groupmembench}, GateMem evaluates utility, access control, and active forgetting in a shared pool~\citep{ren2026gatemem}, and PAC-Bench studies private memories under two-agent privacy constraints~\citep{park2026pacbench}. They show that disclosure matters beyond recall, but scope memory mainly through speakers, identities, roles, or access policies. \comembench instead asks which artifact a consumer requires at its current task state and which plausible artifact must remain outside its context, making sharing and isolation a boundary conditioned by task dependencies, verification, and version.\par}

\subsection{Multi-Agent and Workflow Benchmarks}

{\color{black}AgentBench, MindAgent, and LLM-Coordination evaluate interactive or controlled collaboration~\citep{liu2024agentbench,gong2023mindagent,agashe2025llmcoordination}, and MultiAgentBench varies star, chain, tree, and graph \emph{interaction protocols}~\citep{zhu2025multiagentbench}. TaskBench and MASBench instead vary task dependencies, including chains, DAGs, depth, aggregation, and parallelism~\citep{shen2024taskbench,ke2026masorchestra}. These works establish the importance of organization and workflow structure, but neither interaction topology nor end-to-end success reveals whether memory was selectively exposed according to task dependencies.\par}

{\color{black}$\tau$-bench and AppWorld provide stateful executable outcomes~\citep{yao2025taubench,trivedi2024appworld}; Toolathlon and Toolathlon-GYM provide long-horizon application workflows and reproducible environments~\citep{li2025toolathlon,toolathlongym2026}; and PeopleJoin and SILO-BENCH study distributed information~\citep{jhamtani2025peoplejoin,zhang2026silobench}. These benchmarks ground outcomes or collaboration, but lack controlled tests of required sharing and isolation. \comembench extracts topology-diverse workflows from source task graphs, validates them natively, and tests typed, versioned handoffs with matched clean--polluted runs, separating completion, progress, sharing, and isolation.\par}

\section{The \comembench Benchmark}
\label{sec:benchmark_design}

\begin{figure*}[t]
    \centering
    \includegraphics[width=\textwidth]{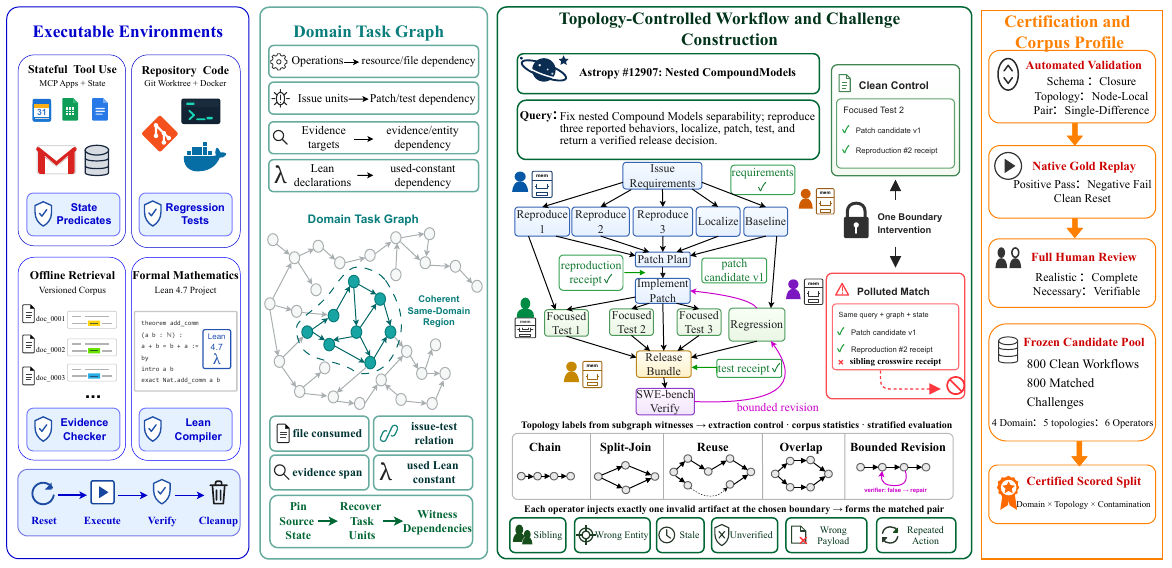}
    \caption{Overview of the \comembench construction pipeline.}
    \label{fig:benchmark-construction}
    \vspace{-1.2em}
\end{figure*}

\comembench represents each instance as an executable task graph with node-local
specifications, typed and version-aware handoffs, resources, and a domain-native
evaluator. Its pipeline recovers domain task graphs, extracts topology-controlled
workflows, adds matched isolation challenges, and certifies their execution.
Figure~\ref{fig:benchmark-construction} summarizes the design.

\subsection{Workflow and boundary \textcolor{black}{operationalization}}
\label{sec:workflow_formulation}

An executable workflow comprises a request $q$, task graph $G=(V,E)$, node-local
specifications $\mathcal{X}=\{x_v\}_{v\in V}$, resources and initial state
$\mathcal{R}$, and a domain-native evaluator $\Phi$:
\begin{equation}
    \mathcal{W}=(q,G,\mathcal{X},\mathcal{R},\Phi).
\end{equation}
{\color{black}Nodes denote executable task units, while edges encode witnessed task dependencies rather than physical agents or an agent communication network. A required edge $e=(u,v)$ carries a handoff contract $h_e=(u,v,\tau,z,\nu,\sigma)$ that records the producer and consumer, semantic type, entity scope, version, and verification requirement. The corresponding artifacts are benchmark-visible workflow records, including evidence, plans, candidate results, execution-state and verification receipts, and domain deliverables.\par}

{\color{black}For a consumer $v$ at workflow state $s$, an information item $m$ is applicable when it can legitimately support $v$ under the current task dependencies and state; otherwise it is inapplicable. The \emph{collaborative memory boundary} is the task- and state-conditioned distinction between these cases. Its required-sharing side preserves information needed by the consumer, while its isolation side prevents inapplicable information from changing the consumer's verified behavior. This notion covers episodic records, structured state, and procedural or skill-like memory, independent of how a system stores or retrieves them.\par}

{\color{black}\comembench makes the two sides observable at declared workflow handoffs. A required artifact provides a positive probe for sharing; a matched, contextually inapplicable artifact provides a negative intervention for isolation. Handoff metadata---including semantic type, entity scope, version, and verification state---supplies the task- and state-specific evidence used to construct and validate these probes. Evaluation then uses artifact exposure and native execution receipts, allowing systems to realize memory through transcripts, retrieval stores, structured representations, or procedural mechanisms.\par}

\subsection{Executable environments}
\label{sec:executable_environments}
\label{sec:testbed}

\comembench executes 800 workflows in domain-native environments. Across 200 workflows per domain, stateful tasks draw from 25 pinned MCP servers detailed in Appendix~\ref{app:mcp_inventory}, activating only the 4--10 declared per task; code tasks span 12 repositories and use issue-specific Docker environments evaluated by the official SWE-bench harness against fixed base commits and \textsc{fail-to-pass}/\textsc{pass-to-pass} tests; retrieval tasks use 200 immutable corpus snapshots containing 16,335 document records, with document-level and exact-span verification; and mathematical tasks compile submitted proofs in isolated copies of three commit-locked Lean~4.7 projects. These native evaluators produce node-level receipts and terminal workflow verdicts.

Each run follows a prepare--execute--verify--cleanup protocol bound to pinned assets, environment configurations, source revisions, and evaluator versions. Stateful tasks additionally record state digests and cleanup receipts, while infrastructure outcomes remain separate from method failures.

\subsection{Recovering domain task graphs}
\label{sec:domain_graph_recovery}

From sources with realistic requirements and executable evaluators, we recover task units and dependencies in four domains: stateful tool use, repository code change, offline retrieval, and formal mathematics. Source instances provide requirements, assets, and evaluators rather than ready-made multi-agent workflows. We pin assets by task identifier, commit, corpus snapshot, or environment digest; shared entities, resources, state transitions, and evaluator contracts then define graph edges.

Recovery follows each source's native structure. We align Toolathlon operations with resources, initialization, and evaluator predicates~\citep{li2025toolathlon,toolathlongym2026}; expand SWE-bench Verified issues into reproduction, localization, patch, and test units~\citep{jimenez2024swebench}; decompose BrowseComp+ items into evidence targets, entity constraints, and synthesis~\citep{chen2025browsecompplus}; and recover declaration graphs and used constants from three miniCTX Lean~4.7 projects~\citep{hu2024minictx}. We exclude declarations transitively depending on \texttt{sorryAx} and retain only edges witnessed by consumed files, issue--test links, evidence requirements, or Lean constants.

\raggedbottom
\subsection{Topology-controlled workflow and challenge construction}
\label{sec:workflow_challenge_construction}

\noindent\textbf{Topology-controlled workflow extraction.}\label{sec:topology_extraction}
Within each domain graph, we select a connected, scenario-coherent subgraph with one terminal objective, complete dependency closure, necessary witnessed edges, and native-evaluator coverage. We assign topology labels from explicit graph witnesses: \emph{chain} denotes a serial dependency; \emph{split--join}, independent branches that reconverge; \emph{reuse}, one typed artifact consumed by multiple downstream regions; \emph{overlap}, consumers with partially shared predecessor sets; and \emph{bounded revision}, one verifier-triggered repair and superseding artifact. Labels are multi-label because a workflow may contain several motifs.

\smallskip
\noindent\textbf{Query and node-local realization.}
\label{sec:query_realization}
{\color{black}For each extracted graph $G_i$, we construct a natural request $q_i$ and a local task $x_v$ for every node. The request specifies the objective, entities, resources, deliverables, constraints, and acceptance conditions without prescribing an execution plan, following Toolathlon's distinction between realistic underspecification and information required for deterministic execution~\citep{li2025toolathlon}. Each node satisfies \emph{node-local closure}: its task, authorized environment view, and declared incoming artifacts contain the information required for execution. Every required handoff records its semantic type, entity scope, version, and verification requirement. Together, node-local closure and declared handoff contracts instantiate the required-sharing side of the collaborative memory boundary: they specify which upstream records each consumer needs and make their delivery and use observable.\par}

\smallskip
\begin{wrapfigure}{r}{0.52\textwidth}
\vspace{-10pt}
\centering
\includegraphics[width=0.96\linewidth]{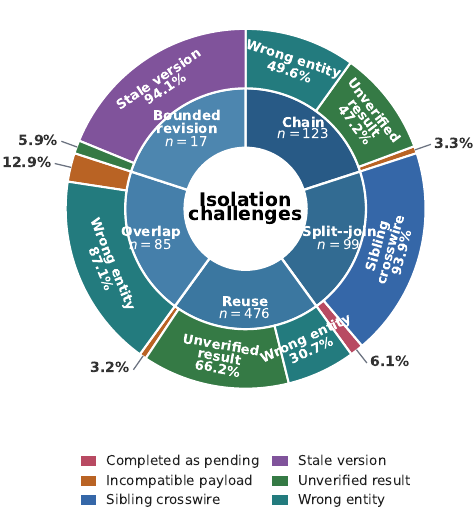}
\captionof{figure}{Topology--operator challenge distribution.}
\label{fig:pollution-topology-composition}
\vspace{-8pt}
\end{wrapfigure}
\noindent\textbf{Matched isolation challenges.}\label{sec:isolation_construction}
{\color{black}For every clean workflow, we select a target consumer supported by a local topology witness and construct a matched variant that inserts exactly one contextually inapplicable artifact into the target consumer's benchmark-visible incoming-artifact view. The challenge label, injection plan, and evaluator predicate remain hidden from the evaluated system. Six operators instantiate concrete violations: \emph{sibling crosswire} routes an artifact from the wrong branch; \emph{incompatible payload} violates the handoff's semantic contract; \emph{unverified result} lacks a required success receipt; \emph{completed-as-pending} presents an already completed side effect as still requiring execution; \emph{wrong entity} binds a type-compatible artifact to a different resource; and \emph{stale version} exposes a superseded artifact. The target topology characterizes the structural setting of the tested handoff, while the operator specifies how the injected artifact becomes inapplicable. Split--joins support crosswires, bounded revisions support stale versions, and chain, reuse, and overlap expose handoff, entity, payload, or verification mismatches; Figure~\ref{fig:pollution-topology-composition} reports the resulting combinations. Each pair otherwise shares its query, graph, environment binding, tools, and non-target inputs. The polluted target is scored only when its clean counterpart verifies successfully. The clean and polluted variants therefore instantiate the two sides of the same tested consumer relation: the clean run establishes that required handoffs support a solvable target, while the matched intervention tests whether one inapplicable artifact changes its verified behavior.\par}

\subsection{Certification and corpus profile}
\label{sec:certification_profile}

\noindent\textbf{Certification.}
A workflow enters the scored split only after three gates. Automated checks enforce source locks, graph closure, acyclicity outside declared revision edges, topology witnesses, node-local closure, control--polluted equality, evaluator coverage, and removal of private material. A private \emph{Gold execution plan} then runs the workflow from a clean reset through the released harness and evaluator; targeted negative mutations must fail, while repeated executions must produce equivalent receipts and semantic state digests. Finally, human reviewers assess query plausibility and completeness, node-local executability, edge necessity, and evaluator--goal alignment. Rejected workflows are repaired and recertified or excluded, and infrastructure defects quarantine the affected family rather than count as method failures. Audit records retain each decision and its receipts, source-grouped splits limit near-duplicate leakage, and Gold plans never enter prompts, public traces, or scores.

\smallskip
\noindent\textbf{Corpus profile.}
The certified release contains 800 clean workflows, evenly distributed across four domains, and 800 matched isolation challenges. Table~\ref{tab:dataset-profile} summarizes per-workflow structural demand and multi-label topology coverage: a workflow contributes every motif present in its graph, whereas receipt-routed conditional revisions are counted as revision boundaries rather than ordinary source-graph edges. 
\begin{table*}[t]
    \centering
    \caption{Structural demand and multi-label topology coverage.}
    \label{tab:dataset-profile}
    \scriptsize
    \setlength{\tabcolsep}{2.7pt}
    \begin{tabular}{lrrrrrrr@{\hspace{5pt}}rrrrrr}
        \toprule
        & \multicolumn{7}{c}{Structural demand (mean)} & \multicolumn{6}{c}{Execution-topology coverage (count)} \\
        \cmidrule(lr){2-8}\cmidrule(lr){9-14}
        Domain & $N$ & Tasks & Depth & Width & Handoffs & Fan-outs & Joins
        & Chain & Split--join & Reuse & Overlap & Revision & Types/WF \\
        \midrule
        Stateful tool use  & 200 & 7.78  & 6.02 & 2.30 & 17.17 & 4.62 & 4.75 & 200 & 200 & 199 & 197 & 70  & 4.33 \\
        Repository code    & 200 & 11.58 & 6.00 & 4.78 & 19.97 & 3.81 & 4.81 & 200 & 200 & 200 & 200 & 74  & 4.37 \\
        Offline retrieval  & 200 & 9.35  & 8.35 & 2.00 & 19.04 & 6.35 & 3.94 & 200 & 200 & 200 & 200 & 126 & 4.63 \\
        Formal mathematics & 200 & 11.70 & 4.59 & 5.58 & 14.47 & 2.88 & 3.56 & 200 & 200 & 200 & 190 & 144 & 4.67 \\
        \midrule
        All / Avg.         & 800 & 10.10 & 6.24 & 3.66 & 17.66 & 4.41 & 4.26 & 800 & 800 & 799 & 787 & 414 & 4.50 \\
        \bottomrule
    \end{tabular}
\end{table*}

Across the release, a workflow averages 10.10 executable task units, 17.66 required handoffs, and 4.50 topology types. The four domains pair these structures with distinct executable artifacts---mutable application state, patches and test evidence, corpus evidence spans, and proof terms with compiler receipts---so the profile captures both graph-level coordination demand and domain-native verification rather than graph size alone.

\section{Evaluation Protocol}
\label{sec:evaluation}

{\color{black}We evaluate workflow execution at two levels. \textbf{SR} measures clean-workflow terminal success, and \textbf{Verified Node Completion Rate} (\VNCR), $|\mathcal{C}|^{-1}\sum_{q\in\mathcal{C}}|\mathcal{V}_q|/|\mathcal{U}_q|$, measures the macro-average verified fraction of required nodes. \textbf{Verified Handoff Success} (\VHS), $|\mathcal{H}^{+}|/|\mathcal{H}|$, is the fraction of eligible consumers that complete successfully after their mandatory direct predecessors verify and their declared artifacts are delivered. \textbf{Isolation Challenge Success} (\ICS), $|\mathcal{P}^{+}|/|\mathcal{P}|$, is polluted-target success among matched pairs whose corresponding clean target succeeds. For the single \ICS value in the system comparison, we set $\mathcal{P}=\bigcup_{d\in\mathcal{D}}\mathcal{P}_d$ and micro-pool eligible matched target pairs across the four domains, so each pair, rather than each domain, receives equal weight. Accordingly, \VHS evaluates the required-sharing side of a collaborative memory boundary, whereas \ICS evaluates its isolation side; SR and \VNCR provide the surrounding task-completion context.\par}

{\color{black}Results are reported by domain and further stratified by target topology and pollution operator. Benchmark and infrastructure failures are separated from method failures, token cost is analyzed in Section~\ref{sec:cost-analysis}, and formal eligibility and accounting rules appear in Appendix~\ref{app:scoring}.\par}

\section{Experiments}
\label{sec:experiments}

{\color{black}We compare systems across domains and backbones, isolate memory mechanisms under a fixed executor, and stratify results by topology and pollution operator to test whether verified progress, required sharing, and isolation improve together.\par}

\subsection{Experimental Setup}
\label{sec:exp-setup}

{\color{black}\noindent\textbf{Evaluated systems.}
We group ten baselines by collaboration abstraction: \emph{general-purpose runtimes}, Codex and Hermes~\citep{openai2026codex,hermes2026}; \emph{multi-agent frameworks}, Deep Agents, CrewAI, and AutoGen~\citep{langchain2026deepagents,crewai2026,wu2024autogen}; \emph{fixed collaboration methods}, Peer Review, SMoA, and OMAC~\citep{xu2023towards,li2025smoa,li2026omac}; and \emph{automated MAS design}, ADAS and EvoMAS~\citep{hu2024automated,hu2026evomas}. Each receives the same node-local task, authorized tools, and declared inputs while retaining its own routing, scheduling, and memory decisions; Appendix~\ref{app:adapters} gives exact configurations.\par}

{\color{black}\noindent\textbf{Controlled memory comparison.}
To isolate memory, a fixed topology-preserving supervisor--worker executor compares \textsc{Local-only}, \textsc{Full Context}, \textsc{Shared Flat RAG}, Mem0~\citep{chhikara2025mem0}, A-Mem~\citep{xu2025mem}, and PlugMem~\citep{yang2026plugmem}. Prompts, workers, tools, graph, evaluator, budgets, and retries remain fixed; Appendix~\ref{app:adapters} specifies each implementation.\par}

{\color{black}\noindent\textbf{Backbones and protocol.}
System comparisons use Qwen3.8-27B, DeepSeek-V4-Flash, and GPT-6-Luna on a fixed domain-stratified split; the memory comparison fixes Qwen3.8-27B. Clean and polluted runs share isolated task-scoped environments and the declared consumer boundary. Manifests record model and framework settings, unsupported combinations are ``--'', and scoring and accounting follow Section~\ref{sec:evaluation} and Appendix~\ref{app:scoring}.\par}

\subsection{System and Backbone Comparison}
\label{sec:exp-overall}

{\color{black}Table~\ref{tab:exp-main-results} tests whether greater verified progress preserves both sides of the boundary. It reports domain-specific SR, \VNCR, and \VHS, while one system-level \ICS micro-pools eligible matched target pairs across domains. All entries are percentages; Section~\ref{sec:evaluation} and Appendix~\ref{app:scoring} define scoring.\par}

\begin{table*}[t]
    \centering
    \caption{System and backbone results.}
    \label{tab:exp-main-results}
    \tiny
    \setlength{\tabcolsep}{1.45pt}
    \renewcommand{\arraystretch}{0.72}
    \resizebox{0.94\textwidth}{!}{%
    \begin{tabular}{@{}cl*{12}{c}c@{}}
        \toprule
        \multirow{2}{*}{Backbone} & \multirow{2}{*}{System} &
        \multicolumn{3}{c}{Stateful Tool} &
        \multicolumn{3}{c}{Repository Code} &
        \multicolumn{3}{c}{Offline Retrieval} &
        \multicolumn{3}{c}{Formal Mathematics} & \multirow{2}{*}{\ICS} \\
        \cmidrule(lr){3-5}\cmidrule(lr){6-8}\cmidrule(lr){9-11}\cmidrule(lr){12-14}
        & & SR & \VNCR & \VHS & SR & \VNCR & \VHS & SR & \VNCR & \VHS & SR & \VNCR & \VHS & \\
        \midrule
        \multirow{10}{*}{Qwen3.8-27B}
        & Codex & 0.0 & 5.1 & 32.7 & 0.0 & 33.2 & 43.0 & 4.1 & 23.3 & 10.7 & 0.0 & 9.1 & 5.3 & 0.0 \\
        & Hermes & 0.0 & 3.2 & 18.2 & 0.0 & 25.2 & 30.2 & 4.1 & 35.7 & 14.4 & 0.0 & 6.2 & 2.0 & 0.0 \\
        \cmidrule(lr){2-15}
        & Deep Agents & 0.0 & 5.5 & 1.4 & 0.0 & 57.7 & 66.8 & 2.0 & 37.8 & 9.5 & 0.0 & 11.4 & 3.4 & 38.1 \\
        & CrewAI & 0.0 & 38.3 & 67.1 & 0.0 & 44.1 & 54.8 & 12.0 & 53.1 & 38.1 & 0.0 & 10.8 & 9.3 & 12.9 \\
        & AutoGen & 0.0 & 33.9 & 56.8 & 0.0 & 48.8 & 55.1 & 18.0 & 50.7 & 37.3 & 0.0 & 10.3 & 8.7 & 20.0 \\
        \cmidrule(lr){2-15}
        & Peer Review & 0.0 & 34.9 & 61.8 & 4.0 & 49.7 & 55.1 & 0.0 & 24.3 & 5.0 & 0.0 & 14.0 & 11.5 & 19.1 \\
        & SMoA & 0.0 & 43.8 & 72.4 & 0.0 & 44.5 & 51.0 & 16.0 & 53.3 & 45.3 & 0.0 & 13.3 & 4.7 & 10.1 \\
        & OMAC & 0.0 & 31.8 & 55.3 & 0.0 & 52.9 & 62.1 & 8.0 & 39.9 & 38.9 & 0.0 & 7.4 & 5.8 & 17.3 \\
        \cmidrule(lr){2-15}
        & ADAS & 0.0 & 18.3 & 36.6 & 0.0 & 17.0 & 37.3 & 0.0 & 12.5 & 0.0 & 0.0 & 6.5 & 8.2 & 7.4 \\
        & EvoMAS & 0.0 & 21.0 & 39.0 & 4.0 & 67.6 & 73.2 & 0.0 & 13.6 & 0.0 & 0.0 & 19.4 & 10.8 & 26.8 \\
        \midrule
        \multirow{10}{*}{DeepSeek-V4-Flash}
        & Codex & 0.0 & 2.7 & 0.0 & 0.0 & 17.1 & 33.2 & 0.0 & 5.1 & 0.0 & 0.0 & 9.5 & 3.9 & 0.0 \\
        & Hermes & 0.0 & 5.6 & 9.4 & 0.0 & 26.5 & 26.1 & 2.0 & 45.7 & 17.1 & 0.0 & 4.2 & 0.0 & 0.0 \\
        \cmidrule(lr){2-15}
        & Deep Agents & 0.0 & 4.3 & 1.8 & 0.0 & 18.8 & 26.1 & 0.0 & 5.3 & 0.0 & 0.0 & 1.4 & 0.0 & 0.0 \\
        & CrewAI & 0.0 & 14.4 & 29.6 & 0.0 & 40.1 & 45.0 & 2.0 & 33.7 & 9.7 & 0.0 & 9.2 & 4.1 & 6.1 \\
        & AutoGen & 0.0 & 15.0 & 28.8 & 0.0 & 63.8 & 71.9 & 4.0 & 40.5 & 24.8 & 0.0 & 10.2 & 6.9 & 15.6 \\
        \cmidrule(lr){2-15}
        & Peer Review & 0.0 & 17.6 & 24.0 & 0.0 & 52.0 & 65.2 & 2.0 & 23.3 & 17.2 & 0.0 & 9.9 & 6.0 & 17.1 \\
        & SMoA & 0.0 & 21.5 & 51.9 & 2.0 & 76.8 & 85.8 & 2.0 & 41.0 & 32.9 & 0.0 & 8.2 & 3.2 & 14.9 \\
        & OMAC & 0.0 & 66.2 & 82.5 & 4.0 & 84.2 & 89.4 & 10.0 & 65.5 & 57.6 & 0.0 & 12.3 & 5.3 & 4.4 \\
        \cmidrule(lr){2-15}
        & ADAS & 0.0 & 17.8 & 52.0 & 8.0 & 67.6 & 80.9 & 4.0 & 29.9 & 16.1 & 0.0 & 11.5 & 2.0 & 14.3 \\
        & EvoMAS & 0.0 & 12.9 & 22.2 & 6.0 & 58.8 & 68.9 & 0.0 & 13.6 & 4.1 & 0.0 & 17.6 & 9.6 & 9.8 \\
        \midrule
        \multirow{10}{*}{GPT-6-Luna}
        & Codex & 0.0 & 6.0 & 2.0 & 0.0 & 13.3 & 17.2 & 0.0 & 3.5 & 0.0 & 0.0 & 4.0 & 0.0 & 0.0 \\
        & Hermes & 0.0 & 6.0 & 43.8 & 0.0 & 25.9 & 25.4 & 0.0 & 31.4 & 11.5 & 0.0 & 14.7 & 0.0 & 0.0 \\
        \cmidrule(lr){2-15}
        & Deep Agents & 0.0 & 3.0 & 0.0 & 0.0 & 12.5 & 32.1 & 0.0 & 8.9 & 0.0 & 0.0 & 0.0 & 0.0 & 0.0 \\
        & CrewAI & 0.0 & 35.7 & 63.1 & 0.0 & 52.5 & 64.4 & 0.0 & 41.6 & 8.2 & 0.0 & 8.4 & 6.0 & 13.3 \\
        & AutoGen & 0.0 & 19.0 & 69.2 & 0.0 & 37.6 & 73.3 & 4.0 & 41.6 & 31.1 & 0.0 & 12.0 & 2.4 & 11.8 \\
        \cmidrule(lr){2-15}
        & Peer Review & 0.0 & 13.7 & 35.8 & 0.0 & 3.1 & 20.0 & 0.0 & 10.2 & 0.0 & 0.0 & 4.0 & 0.0 & 0.0 \\
        & SMoA & 0.0 & 29.6 & 68.5 & 0.0 & 49.2 & 64.1 & 2.0 & 43.5 & 17.0 & 0.0 & 11.2 & 2.5 & 15.1 \\
        & OMAC & 0.0 & 39.7 & 66.9 & 0.0 & 35.3 & 72.4 & 0.0 & 28.5 & 12.5 & 0.0 & 7.8 & 1.4 & 4.4 \\
        \cmidrule(lr){2-15}
        & ADAS & 0.0 & 26.8 & 69.8 & 0.0 & 26.7 & 31.3 & 2.0 & 36.0 & 9.4 & 0.0 & 6.5 & 7.4 & 7.1 \\
        & EvoMAS & 0.0 & 43.6 & 63.7 & 0.0 & 55.5 & 67.9 & 0.0 & 17.0 & 5.2 & 0.0 & 18.0 & 16.4 & 7.0 \\
        \bottomrule
    \end{tabular}%
    }
\end{table*}

\paragraph{Task success and verified progress.}
{\color{black}No supported configuration solves a Stateful Tool or Formal Mathematics workflow, and the best SR is only 8.0 in Repository Code and 18.0 in Offline Retrieval. Yet some zero-SR configurations reach 84.2 \VNCR and 89.4 \VHS, verifying substantial subgraphs and downstream consumers without closing the terminal objective. The \VNCR--\VHS gap further shows that upstream completion does not guarantee a successful required handoff; workflow completion, verified progress, and sharing reliability expose distinct failure stages.\par}

\paragraph{Sharing versus isolation.}
{\color{black}Required sharing and isolation do not improve together. General-purpose runtimes are generally weak in verified progress, while explicit multi-agent organizations improve particular domains without dominating both boundary requirements. With Qwen3.8-27B, SMoA reaches 43.8/72.4 \VNCR/\VHS on Stateful Tool but only 10.1 \ICS; Deep Agents instead attains 38.1 \ICS despite weak progress. With DeepSeek-V4-Flash, OMAC leads progress in Stateful Tool, Code, and Retrieval yet reaches only 4.4 \ICS, below Peer Review's 17.1; rankings shift again under GPT-6-Luna. Because \VHS tests required handoffs and \ICS conditions on a successful clean target, stronger execution and coordination do not by themselves preserve both sides of the boundary.\par}

\subsection{Memory Mechanism Comparison}
\label{sec:exp-memory}

{\color{black}Table~\ref{tab:exp-memory-results} fixes the executor, workers, handoffs, tools, and evaluator, varying only additional cross-worker exposure. It retains domain-specific \ICS because memory policies may interact with domain-native artifacts. Relative to \textsc{Local-only}, which preserves declared handoffs but disables extra cross-worker memory, \textsc{Full Context} raises Stateful Tool \VNCR from 74.6 to 86.9 and Code SR from 0.0 to 16.0, but lowers \ICS from 97.1 to 72.4 and from 89.4 to 71.9. \textsc{Shared Flat RAG} reaches 90.8 Code \VHS yet only 73.9 \ICS, showing that broader or semantically relevant exposure can improve availability while weakening task- and state-dependent isolation.\par}

{\color{black}Structured stores likewise do not dominate all outcomes. A-Mem reaches 18.4 Code SR but only 63.9 \ICS, whereas PlugMem preserves 91.3 Code \ICS with 0.0 SR and falls to 52.9 \ICS on Retrieval. Formal Mathematics remains unsolved at workflow level. Thus retention, linking, and procedural organization may help particular domains without consistently preserving both required access and isolation.\par}

\begin{table}[!t]
    \centering
    \caption{Memory-mechanism results under controlled multi-agent execution.}
    \label{tab:exp-memory-results}
    \tiny
    \setlength{\tabcolsep}{1.25pt}
    \renewcommand{\arraystretch}{0.88}
    \resizebox{\textwidth}{!}{%
    \begin{tabular}{@{}l*{16}{r}@{}}
        \toprule
        \multirow{2}{*}{Memory mechanism} &
        \multicolumn{4}{c}{Stateful Tool} &
        \multicolumn{4}{c}{Repository Code} &
        \multicolumn{4}{c}{Offline Retrieval} &
        \multicolumn{4}{c}{Formal Mathematics} \\
        \cmidrule(lr){2-5}\cmidrule(lr){6-9}\cmidrule(lr){10-13}\cmidrule(lr){14-17}
        & SR & \VNCR & \VHS & \ICS & SR & \VNCR & \VHS & \ICS & SR & \VNCR & \VHS & \ICS & SR & \VNCR & \VHS & \ICS \\
        \midrule
        \multicolumn{17}{@{}l}{\textit{Controls}} \\
        \textsc{Local-only}       & 0.0 & 74.6 & 84.0 & 97.1 & 0.0 & 77.1 & 89.5 & 89.4 & 2.0 & 63.6 & 83.9 & 60.9 & 0.0 & 13.6 & 26.9 & 0.0 \\
        \textsc{Full Context}     & 2.0 & 86.9 & 86.7 & 72.4 & 16.0 & 61.7 & 89.1 & 71.9 & 6.0 & 77.1 & 88.6 & 80.6 & 0.0 & 11.8 & 26.8 & 0.0 \\
        \textsc{Shared Flat RAG}  & 2.0 & 86.9 & 86.4 & 70.6 & 18.0 & 72.4 & 90.8 & 73.9 & 8.0 & 74.1 & 88.4 & 75.0 & 0.0 & 14.4 & 29.7 & 20.0 \\
        \addlinespace[1pt]
        \multicolumn{17}{@{}l}{\textit{External memory systems}} \\
        \textsc{Mem0}             & 0.0 & 83.8 & 85.5 & 86.3 & 2.0 & 77.9 & 89.6 & 89.1 & 4.0 & 64.9 & 84.7 & 73.9 & 0.0 & 12.2 & 22.7 & 52.4 \\
        \textsc{A-Mem}            & 4.0 & 85.8 & 86.6 & 87.9 & 18.4 & 58.2 & 88.4 & 63.9 & 6.0 & 71.8 & 85.6 & 71.9 & 0.0 & 12.3 & 26.2 & 0.0 \\
        \textsc{PlugMem}          & 0.0 & 83.1 & 85.1 & 90.0 & 0.0 & 68.2 & 88.9 & 91.3 & 4.0 & 56.9 & 83.9 & 52.9 & 0.0 & 14.2 & 29.0 & 33.3 \\
        \bottomrule
    \end{tabular}%
    }
\end{table}

\subsection{Topology and Isolation Analysis}
\label{sec:exp-structure}

{\color{black}We stratify Qwen3.8-27B traces by target topology in Table~\ref{tab:exp-topology} and pollution operator in Appendix Table~\ref{tab:exp-pollution}; both summarize Table~\ref{tab:exp-main-results}. Topology identifies where a consumer boundary is tested, while the operator identifies why an artifact is inapplicable. Within each cross-domain stratum, \VNCR pools canonical nodes, \VHS eligible consumers, and \ICS matched pairs whose clean target succeeds; workflows are not assigned wholesale to every motif. All entries are percentages.\par}

\begin{table}[t]
    \centering
    \caption{Topology-stratified task progress and isolation success.}
    \label{tab:exp-topology}
    \scriptsize
    \setlength{\tabcolsep}{1.8pt}
    \renewcommand{\arraystretch}{1.02}
    \resizebox{\textwidth}{!}{%
    \begin{tabular}{@{}l*{15}{c}@{}}
        \toprule
        \multirow{2}{*}{System} &
        \multicolumn{3}{c}{Chain} &
        \multicolumn{3}{c}{Split--join} &
        \multicolumn{3}{c}{Reuse} &
        \multicolumn{3}{c}{Overlap} &
        \multicolumn{3}{c}{Bounded revision} \\
        \cmidrule(lr){2-4}\cmidrule(lr){5-7}\cmidrule(lr){8-10}\cmidrule(lr){11-13}\cmidrule(lr){14-16}
        & \VNCR & \VHS & \ICS & \VNCR & \VHS & \ICS & \VNCR & \VHS & \ICS & \VNCR & \VHS & \ICS & \VNCR & \VHS & \ICS \\
        \midrule
        Codex & 29.7 & 62.3 & 0.0 & 20.2 & 29.1 & 0.0 & 22.8 & 25.9 & 0.0 & 21.9 & 29.6 & 0.0 & 20.7 & 31.2 & 0.0 \\
        Hermes & 28.4 & 56.5 & 0.0 & 20.1 & 20.5 & -- & 22.7 & 14.8 & 0.0 & 21.4 & 19.8 & -- & 25.0 & 0.0 & -- \\
        \cmidrule(lr){1-16}
        Deep Agents & 42.4 & 81.3 & 33.3 & 33.4 & 40.1 & 42.9 & 38.3 & 39.2 & 28.6 & 36.1 & 40.4 & 80.0 & 32.5 & 30.8 & -- \\
        CrewAI & 45.7 & 81.1 & 20.0 & 40.0 & 49.7 & 0.0 & 47.3 & 49.6 & 16.2 & 42.8 & 50.0 & 0.0 & 43.3 & 47.5 & 0.0 \\
        AutoGen & \textbf{47.9} & 80.9 & 33.3 & 39.9 & 46.8 & 25.0 & 46.9 & 46.6 & 19.4 & 42.6 & 47.0 & 0.0 & 43.3 & 52.2 & 0.0 \\
        \cmidrule(lr){1-16}
        Peer Review & 46.6 & \textbf{85.1} & 44.4 & 32.5 & 43.1 & 0.0 & 36.5 & 43.0 & 14.3 & 34.9 & 43.9 & 0.0 & 28.1 & 53.4 & 20.0 \\
        SMoA & 47.0 & 79.5 & 25.0 & \textbf{41.8} & 51.7 & 0.0 & \textbf{49.4} & 53.8 & 11.9 & \textbf{44.1} & 52.2 & 0.0 & \textbf{44.6} & \textbf{55.0} & 0.0 \\
        OMAC & 45.3 & 84.4 & 28.6 & 36.5 & 51.0 & 16.7 & 43.3 & 50.3 & 13.3 & 39.1 & 51.1 & 25.0 & 33.9 & 38.1 & 20.0 \\
        \cmidrule(lr){1-16}
        ADAS & 16.5 & 56.7 & 0.0 & 14.2 & 26.9 & 33.3 & 16.4 & 25.3 & 0.0 & 15.4 & 27.7 & 0.0 & 14.5 & 42.6 & 0.0 \\
        EvoMAS & 46.6 & 79.3 & 30.0 & 33.8 & \textbf{51.8} & 72.7 & 37.3 & \textbf{54.8} & 14.3 & 36.1 & \textbf{52.8} & 0.0 & 20.3 & 26.7 & 0.0 \\
        \bottomrule
    \end{tabular}
    }
\end{table}

{\color{black}Boundary maintenance is not a single system-level capability. Chain consumers have the highest \VHS for every system, while branch separation, reuse, overlap, and revision change which records must be routed, separated, or superseded. Rankings do not follow one difficulty order: EvoMAS reaches 72.7 \ICS on split--join but 14.3 on reuse, whereas Deep Agents reaches 80.0 on overlap but 28.6 on reuse. Protecting one local boundary therefore does not ensure robustness when information is reused or partially shared.\par}

{\color{black}Appendix Table~\ref{tab:exp-pollution} provides the complementary semantic diagnosis: robustness to one violation does not reliably transfer to another. Topology determines where an artifact can affect execution, whereas semantics, scope, verification, and version determine whether it should affect that consumer. These strata explain complementary boundary pressures but are descriptive rather than causal topology--operator comparisons.\par}
\begin{figure*}[t]
	\vspace{-0.65em}
	\centering
	\includegraphics[width=0.93\textwidth]{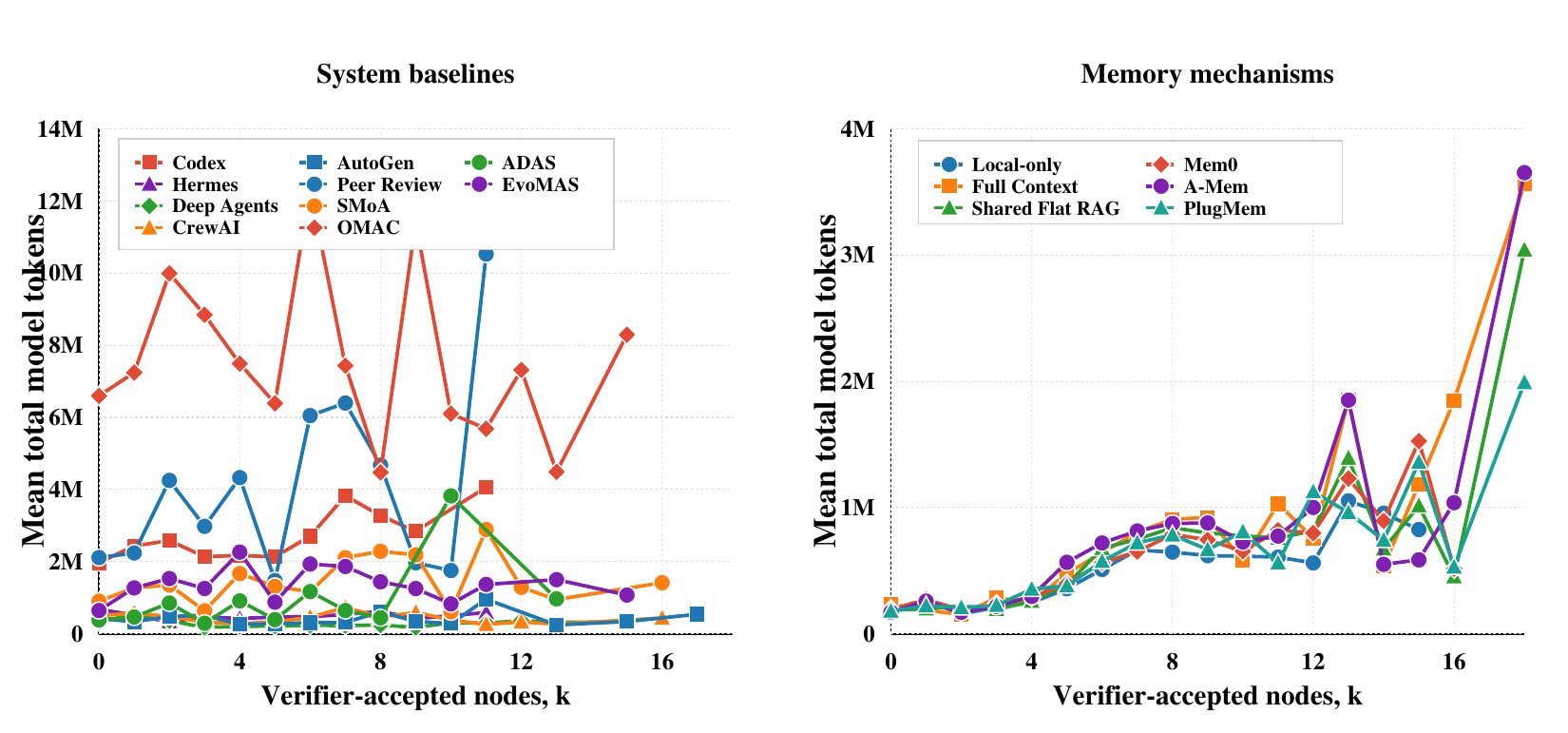}
	\caption{\textcolor{black}{Progress-conditioned token cost; panels use separate linear scales.}}
	\label{fig:progress-conditioned-cost}
	\vspace{-0.5em}
\end{figure*}

\subsection{Cost Analysis}
\label{sec:cost-analysis}

{\color{black}Figure~\ref{fig:progress-conditioned-cost} compares tokens at equal verifier-accepted progress. Deep Agents' supervisor--subagent loop averages 0.31M tokens per task, versus 8.01M for OMAC's multi-round process and 3.61M for Peer Review's create--review--revise loop. None dominates SR, \VHS, or \ICS, so coordination cost guarantees neither reliable handoffs nor isolation (Appendix~\ref{app:scoring}).\par}

{\color{black}Compared with the 489k-token \textsc{Local-only} control, \textsc{Full Context} and \textsc{Shared Flat RAG} use 672k and 621k tokens without consistently improving both sharing and isolation; PlugMem costs 521k but varies by domain. Efficient collaborative memory therefore requires selective exposure, not simply more context.\par}

\FloatBarrier

\section{Conclusion}
\label{sec:conclusion}

{\color{black}Multi-agent workflows must reuse upstream information without allowing contextually inapplicable records to alter downstream execution; topology makes this boundary vary across consumers and states. \comembench operationalizes this problem with 800 executable workflows spanning four domains and diverse topologies. Declared handoffs measure required sharing, while matched challenges and native evaluators test isolation through downstream behavior. Across systems, backbones, and memory mechanisms, completion, verified progress, handoff reliability, and isolation remain distinct: broader access can improve progress without preserving isolation, and rankings vary by topology and violation. Effective collaborative memory therefore requires selective, task- and state-conditioned exposure rather than simply more context.}

\subsection*{AI Use Statement}
{\color{black}Generative AI tools were used for language and presentation editing and
routine implementation assistance. The authors designed the study, verified the data
and results, and take full responsibility for the paper and released artifacts.}

\subsection*{Reproducibility Statement}
We will release the benchmark data, clean--polluted pair specifications, schemas,
evaluation code, prompts, and environment configurations. Each run records system and
model versions, decoding and budget settings, source and evaluator revisions, and asset
checksums. Appendix~\ref{app:schema} documents the released records,
Appendix~\ref{app:construction} describes construction and certification, and
Appendices~\ref{app:evaluators}--\ref{app:scoring} specify the execution substrates,
baseline implementations, and scoring protocol.

\bibliography{references,benchmark_references}
\bibliographystyle{iclr2027_conference}

\appendix
\section{Benchmark Schema and Runtime Protocol}
\label{app:schema}

The release separates reusable workflow definitions from executions. A
\texttt{workflow\_family} record stores the source references, global request,
node-local tasks, typed dependency edges, topology witnesses, execution contract,
and split. A \texttt{benchmark\_instance} binds one family to either its clean
control or matched polluted variant; the two variants share the family, scenario,
and environment initialization. The public package contains 800 families, 1,600
instances, and 8,975 node-local contexts. Private evaluator oracles and pollution
plans are distributed only with the operator package.

At execution time, a worker receives a \texttt{node\_view}: its role and task,
authorized local context, declared incoming artifacts, handoff contract, output
type, and an opaque capability token. An artifact records its semantic type,
producer, scenario and entity scope, version, verification state, payload, and
provenance. The native evaluator returns a verification receipt, and only a verified
receipt permits a required artifact to satisfy a downstream dependency. Systems
that control routing may emit a \texttt{memory\_proposal} containing an artifact,
producer, and proposed recipients; the harness records the proposal and the
artifacts actually exposed to each consumer.

These records describe the benchmark-visible boundary, not an architecture's
internal memory. A system may use a transcript, vector store, structured memory,
or no persistent store, but private reasoning and store contents are neither
inspected nor treated as ground truth. Replay instead follows the manifest-fixed
case initialization, node views, native actions, receipts, issued artifacts,
handoffs, consumed-artifact identifiers, and pollution events. Schema validation,
content hashes, and receipt signatures make an exposure trace replayable without
assuming a common internal memory representation.

\section{Data Sources and Construction Details}
\label{app:construction}

Table~\ref{tab:app-source-construction} records the pinned sources and the evidence
used to recover task dependencies. Source repositories are frozen at the listed
revisions; their license and attribution files are preserved. Toolathlon-GYM is
Apache-2.0, SWE-bench and BrowseComp+ are MIT licensed, and the three Lean projects
retain their project-specific upstream licenses. MCP-server components remain
subject to their own licenses rather than being relicensed by \comembench.

\begin{table*}[t]
    \centering
    \scriptsize
    \setlength{\tabcolsep}{3.2pt}
    \renewcommand{\arraystretch}{1.04}
    \caption{Pinned sources and domain-graph recovery.}
    \label{tab:app-source-construction}
    \begin{tabularx}{\textwidth}{@{}p{0.13\textwidth}p{0.23\textwidth}p{0.27\textwidth}X@{}}
        \toprule
        Domain & Source snapshot & Task units & Dependency evidence \\
        \midrule
        Stateful tool use & Toolathlon-GYM \texttt{45bb4935} & Source-authored operations over named resources and side effects & Resource flow, state transitions, evaluator predicates, and typed source-workflow dependencies \\
        Repository code & SWE-bench Verified \texttt{c104f84} & Issue analysis, reproduction, localization, patching, and regression checks for one issue & Base commit, issue--test links, modified symbols, and \textsc{fail-to-pass}/\textsc{pass-to-pass} contracts \\
        Offline retrieval & BrowseComp+ \texttt{144cff8e} & Query constraints, fixed-corpus evidence partitions, entity resolution, answer evidence, and synthesis & Hidden qrels, stable document identifiers, exact spans, constraint coverage, and provenance requirements \\
        Formal mathematics & HepLean \texttt{7448822a}; HTPI \texttt{8eeebaec}; PNT \texttt{23650db8} & Lean declarations in an elaborated target cone & Proof-term used constants, source modules and lines, and Lean~4.7 elaboration \\
        \bottomrule
    \end{tabularx}
\end{table*}

For each source record, we first build a domain-level graph and then extract a
connected subgraph with one terminal objective, complete ancestor closure, and a
native evaluator. No workflow joins unrelated source cases or crosses domains.
Selection retains only edges with the evidence in Table~\ref{tab:app-source-construction}
and favors subgraphs that instantiate the released topology motifs without adding
synthetic dependencies. The request is then realized from the selected source
goal, entities, resources, deliverables, constraints, and acceptance conditions;
node tasks expose only the local information and declared predecessors required to
perform their work. This produces a new multi-agent workflow while preserving the
source task and its executable semantics.

Splits are assigned at family level, so the clean and polluted variants and all
node views of a workflow remain together. Repository and Lean-project group checks
additionally prevent the same repository or proof project from crossing splits.
The released assignment contains 177 development and 623 test families, balanced
at 200 families per domain overall.

\section{Quality Assurance and Audit Protocol}
\label{app:audit}

Certification combines exhaustive machine gates with semantic review. JSON Schema
validation covers families, instances, contexts, node views, artifacts, proposals,
and private oracles. Graph gates check identifier closure, reachable terminal
objectives, acyclicity outside declared revision edges, topology witnesses, and a
dependency-evidence record on every edge. Node-local closure is checked against the
task, authorized resources and capabilities, and required incoming handoffs; an
undeclared path, tool, predecessor, or output semantic fails the gate. All 800
families pass the released graph and execution-contract audits.

Evaluator closure is tested by replay rather than final-text judging. The
deterministic graph-and-handoff oracle accepts the known-good trace and rejects its
targeted negative trace for all 800 families. Domain-native evaluators then determine
node and terminal success during benchmark execution; all 200 retrieval families
additionally pass document, qrel, and exact-span closure. For isolation challenges, every plan
must identify a real donor, a target handoff, an invalidity reason, a local topology
witness, and a deterministic observable failure predicate. The candidate gate passes
for all 800 matched plans, and its single-difference manifest certifies that query,
graph, evaluator, environment, and non-target inputs are unchanged.

Human review is repair-oriented rather than a majority-vote annotation task.
Reviewers compare each candidate with its source record and native evaluator using
four questions: whether the request is natural and complete, whether every node is
executable from its local view, whether each retained edge is necessary for its
consumer, and whether evaluator acceptance entails the requested outcome. A disputed
item is adjudicated by rechecking the source evidence and evaluator contract; it is
then repaired and rerun through all gates or excluded. We therefore do not treat an
automatic language-model score as evidence of realism or report an artificial
inter-annotator agreement statistic.

Finally, a public--private leakage audit rejects any public occurrence of Gold
answers or patches, qrel mappings, pollution operators, target labels, failure
predicates, or evaluator-only fields. Source-group split checks prevent repository
and proof-project leakage, and digest comparisons detect reset drift. The final
release audit reports no schema, graph, oracle-replay, source-invariant, or public
leakage findings.

\section{Native Evaluators and Environment Setup}
\label{app:evaluators}

\subsection{MCP application environment}
\label{app:mcp_inventory}

The application environment comprises 25 MCP servers. MCP defines the agent-facing action protocol, while the corresponding resources reside in task workspaces, PostgreSQL-backed applications, and locally managed services. Before a workflow begins, \comembench materializes its initial files and application state in a family-isolated workspace, database, and service namespace. At each node, the harness starts only the servers admitted by the node capability contract. It checks process startup and tool-schema discovery before exposing the tools, and records the arguments, native observations, and state changes of executed calls. Environment initialization, state export, evaluator invocation, and cleanup remain benchmark control operations and are not available to evaluated agents.

\begin{table}[H]
    \centering
    \footnotesize
    \setlength{\tabcolsep}{4pt}
    \renewcommand{\arraystretch}{1.12}
    \caption{MCP server inventory in the pinned application-task source environment. Runtime identifiers are normalized to readable server names.}
    \label{tab:mcp-inventory}
    \begin{tabularx}{\linewidth}{@{}>{\raggedright\arraybackslash}p{0.23\linewidth}>{\raggedright\arraybackslash}Xr@{}}
        \toprule
        \textbf{Capability group} & \textbf{MCP servers} & \textbf{Count} \\
        \midrule
        Files and office documents & Filesystem, Excel, Word, PowerPoint, PDF Tools & 5 \\
        Collaboration and productivity & Email, Google Calendar, Google Forms, Google Sheets, Notion, Canvas, Memory & 7 \\
        Research and web content & Local arXiv, arXiv LaTeX, Scholarly Search, Fetch, Playwright, YouTube, YouTube Transcript, HowToCook & 8 \\
        Business and structured data & Snowflake, WooCommerce, Yahoo Finance, 12306 Rail & 4 \\
        Local computation & Terminal & 1 \\
        \midrule
        Total & & 25 \\
        \bottomrule
    \end{tabularx}
\end{table}

Server source, dependencies, build artifacts, and launch configurations are pinned before evaluation and applied uniformly to all systems. Environment certification records server entrypoints, database routing, initial and final state digests, and cleanup receipts; a task with an unavailable declared server is excluded from method-level scoring.

\subsection{Non-MCP native substrates}

The other execution substrates retain their native interfaces instead of being artificially wrapped as MCP services. Code outputs are submitted to the official SWE-bench evaluation entrypoint, \texttt{swebench.harness.run\_evaluation}, which applies the patch inside the instance-specific Docker environment and executes its FAIL\_TO\_PASS and PASS\_TO\_PASS tests. Mathematical outputs are inserted into an isolated copy of the corresponding Lean source file and checked with the pinned Lake-managed Lean~4.7 toolchain; the original source is restored after compilation. Retrieval nodes operate over immutable corpus snapshots and submit an answer with document identifiers and exact source spans. The retrieval verifier checks document existence, span provenance, declared constraint coverage, and answer correctness; it evaluates evidence selection and synthesis rather than open-web search behavior. All three substrates emit the same canonical action, artifact, receipt, and failure records used by the MCP path.

\section{Baseline Instantiations and Run Manifests}
\label{app:adapters}

Table~\ref{tab:baseline-instantiations} records the exact implementation evaluated for each
system. The adapter supplies node-local tasks, authorized actions, and declared incoming
artifacts, and records traces without choosing answers, routing, schedules, or memory contents.

\begin{table}[H]
    \centering
    \scriptsize
    \setlength{\tabcolsep}{3pt}
    \renewcommand{\arraystretch}{1.03}
    \caption{System baseline implementations.}
    \label{tab:baseline-instantiations}
    \begin{tabularx}{\linewidth}{@{}>{\raggedright\arraybackslash}p{0.16\linewidth}>{\raggedright\arraybackslash}p{0.20\linewidth}>{\raggedright\arraybackslash}X@{}}
        \toprule
        \textbf{Baseline} & \textbf{Version / revision} & \textbf{Evaluated instantiation} \\
        \midrule
        Codex & CLI 0.152.1 & Official \texttt{codex exec --json} loop with isolated configuration and native event capture. \\
        Hermes & 0.21.0 (\texttt{d3e2ace}) & Native \texttt{AIAgent.chat} and \texttt{delegate\_task} execution with sandbox tools. \\
        \cmidrule(lr){1-3}
        Deep Agents & 0.7.12 (\texttt{3db758c}) & Official \texttt{create\_deep\_agent} supervisor--subagent loop with native model and sandbox-tool bindings. \\
        CrewAI & 1.15.18 (\texttt{b608a35}) & Native \texttt{Agent}/\texttt{Task}/\texttt{Crew} objects with sequential process orchestration. \\
        AutoGen & 0.7.5 (\texttt{027ecf0}) & \texttt{RoundRobinGroupChat} with native messages and tool callbacks. \\
        \cmidrule(lr){1-3}
        Peer Review & EvoMAS source \texttt{93fd9d6} & Three creators produce candidates, followed by nine reviews, three revisions, and aggregation. \\
        SMoA & Source \texttt{1f3444d} & Official sparse-mixture process with its role layers and aggregation policy preserved. \\
        OMAC & Source \texttt{6bccfec} & Frozen five-dimensional, three-round LLMLP package; optimized functions and collaboration structure are retained. \\
        \cmidrule(lr){1-3}
        ADAS & Source \texttt{2702bee} & The selected agent program is frozen before benchmark evaluation and executed through the official agent interface. \\
        EvoMAS & Source \texttt{93fd9d6}; smolagents 1.26.0 & Frozen evolved role, prompt, and organization configuration executed by \texttt{MasRuntime}. \\
        \bottomrule
    \end{tabularx}
\end{table}

\paragraph{Controlled memory mechanisms.}
All six mechanisms run in the same supervisor--worker executor. \textsc{Local-only} exposes
only worker-local state; \textsc{Full Context} exposes every prior visible artifact without
retrieval; and \textsc{Shared Flat RAG} retrieves the top four artifacts from one workflow-local
store using \texttt{all-MiniLM-L6-v2}. Mem0 0.1.104 uses its local \texttt{add}/\texttt{search}
API with a workflow-specific Qdrant store. A-Mem (source \texttt{0c8039f}) writes verified
artifacts as linked notes and retrieves related notes for each node. PlugMem (source
\texttt{3b2ce75}) uses the official graph write, recall, and reasoning path. Each external
store is empty at workflow start; only the memory policy changes across rows.

Every run has an immutable manifest recording source and adapter revisions, model and decoding
settings, budgets, configuration hashes, memory policy, data split, environment and evaluator
hashes, seed, and trace schema. This separates a baseline's organization from the common
execution and scoring substrate.

\section{Metric Definitions and Scoring Details}
\label{app:scoring}

For clean controls $\mathcal{C}$, SR is $|\mathcal{C}|^{-1}\sum_{q\in\mathcal{C}}
\mathbb{I}[\mathrm{terminal\_success}(q)]$. Let $\mathcal{U}_q$ be the canonical required
nodes of $q$ and $\mathcal{V}_q\subseteq\mathcal{U}_q$ those with a verified completion;
\VNCR is $|\mathcal{C}|^{-1}\sum_{q\in\mathcal{C}}|\mathcal{V}_q|/|\mathcal{U}_q|$.
Repeated executions and revision retries count once, while method-private nodes never enter
either set.

For \VHS, let $\mathcal{H}$ be non-root consumers whose mandatory direct predecessors have
verified successfully, and let $\mathcal{H}^{+}$ be the successful subset. A consumer enters
$\mathcal{H}^{+}$ only if every required predecessor artifact is exposed at the declared
boundary, current, receipt-valid, explicitly consumed, and followed by native-verifier success.
Thus \VHS is $|\mathcal{H}^{+}|/|\mathcal{H}|$; unsupported slices are marked ``--''.

Each isolation pair $z$ declares one target consumer $t_z$. Let $\mathcal{P}$ contain pairs for
which $t_z$ succeeds in the clean variant, and $\mathcal{P}^{+}$ those for which it also succeeds
in the matched polluted variant. Thus \ICS is $|\mathcal{P}^{+}|/|\mathcal{P}|$; eligibility is
not terminal success of the full clean workflow. \textcolor{black}{For the pooled system-level score,
$\mathcal{P}=\bigcup_{d\in\mathcal{D}}\mathcal{P}_d$ is the union of eligible pair sets across
domains; domain-specific scores restrict $\mathcal{P}$ to the eligible pair set of one domain.}
Clean and polluted variants share the query,
graph, initial state, tools, and non-target inputs; only one evaluator-hidden inapplicable
artifact at $t_z$ differs.

Token accounting includes execution, coordination, memory-retrieval context, recovery or
repair, and meta-level search or evolution. We report input/output tokens, mean, median, 90th
percentile, successful-run mean, and total valid-run tokens divided by successful workflows.
Benchmark-visible artifact exposure is reported separately. \textcolor{black}{The eligibility conditions for \VHS and \ICS follow the definitions in Section~\ref{sec:evaluation}.} Benchmark and infrastructure failures are excluded from
method-failure denominators.

\section{Additional Evaluation Details}
\label{app:additional_results}

\paragraph{Pollution-operator breakdown.}
{\color{black}Table~\ref{tab:exp-pollution} stratifies the Qwen3.8-27B system runs by the violation that makes the injected artifact inapplicable. Eligible matched pairs are pooled across domains within each operator. EvoMAS is comparatively robust to sibling crosswires but remains vulnerable to unverified results and other observed violations, while Deep Agents performs strongly on wrong-entity challenges but less reliably on unverified results. The variation across columns shows that isolation robustness does not transfer uniformly between artifact violations.\par}

\begin{table}[H]
    \centering
    \caption{Isolation by pollution operator.}
    \label{tab:exp-pollution}
    \scriptsize
    \setlength{\tabcolsep}{2.6pt}
    \renewcommand{\arraystretch}{0.92}
    \resizebox{0.74\textwidth}{!}{%
    \begin{tabular}{@{}l*{6}{c}@{}}
        \toprule
        System & \makecell{Sibling\\crosswire} & \makecell{Incompatible\\payload} & \makecell{Unverified\\result} & \makecell{Completed as\\pending} & \makecell{Wrong\\entity} & \makecell{Stale\\version} \\
        \midrule
        Codex & 0.0 & 0.0 & 0.0 & -- & 0.0 & 0.0 \\
        Hermes & -- & -- & 0.0 & -- & -- & -- \\
        \cmidrule(lr){1-7}
        Deep Agents & 42.9 & 0.0 & 31.0 & -- & 80.0 & -- \\
        CrewAI & 0.0 & 0.0 & 22.2 & 0.0 & 0.0 & 0.0 \\
        AutoGen & 33.3 & 0.0 & 26.3 & 0.0 & 0.0 & 0.0 \\
        \cmidrule(lr){1-7}
        Peer Review & 0.0 & 0.0 & 29.6 & 0.0 & 0.0 & 20.0 \\
        SMoA & 0.0 & 0.0 & 17.5 & 0.0 & 0.0 & 0.0 \\
        OMAC & 33.3 & 0.0 & 20.7 & 0.0 & 12.5 & 20.0 \\
        \cmidrule(lr){1-7}
        ADAS & 33.3 & 0.0 & 0.0 & -- & 0.0 & 0.0 \\
        EvoMAS & 80.0 & 16.7 & 20.7 & 0.0 & 0.0 & 0.0 \\
        \bottomrule
    \end{tabular}%
    }
\end{table}

Table~\ref{tab:app-token-accounting} reports the aggregate token statistics behind
Figure~\ref{fig:progress-conditioned-cost}. Values include model calls used for
execution, coordination, recovery, and memory operations and are computed over runs
with complete usage telemetry.

\begin{table}[H]
    \centering
    \scriptsize
    \setlength{\tabcolsep}{4.0pt}
    \renewcommand{\arraystretch}{0.95}
    \caption{Token accounting in thousands of tokens.}
    \label{tab:app-token-accounting}
    \begin{tabular}{@{}lrrr@{\hspace{18pt}}lrrr@{}}
        \toprule
        System & Mean & Median & Per verified node & Memory mechanism & Mean & Median & Per verified node \\
        \midrule
        Codex & 2350 & 1875 & 1098 & Local-only & 489 & 367 & 80 \\
        Hermes & 518 & 382 & 194 & Full Context & 672 & 380 & 104 \\
        \cmidrule(lr){1-4}
        Deep Agents & 310 & 227 & 89 & Shared Flat RAG & 621 & 389 & 94 \\
        CrewAI & 461 & 263 & 112 & Mem0 & 576 & 422 & 91 \\
        AutoGen & 388 & 231 & 97 & A-Mem & 617 & 357 & 102 \\
        \cmidrule(lr){1-4}
        Peer Review & 3608 & 2265 & 1022 & PlugMem & 521 & 392 & 90 \\
        SMoA & 1419 & 858 & 333 & & & & \\
        OMAC & 8012 & 5536 & 1981 & & & & \\
        \cmidrule(lr){1-4}
        ADAS & 565 & 436 & 290 & & & & \\
        EvoMAS & 1306 & 977 & 343 & & & & \\
        \bottomrule
    \end{tabular}
\end{table}

Table~\ref{tab:app-failure-attribution} states the failure attribution applied
before metric aggregation. Only method failures enter method-performance
denominators; benchmark and infrastructure failures remain separately auditable.

\begin{table}[H]
    \centering
    \scriptsize
    \setlength{\tabcolsep}{3.5pt}
    \renewcommand{\arraystretch}{1.03}
    \caption{Run-outcome attribution.}
    \label{tab:app-failure-attribution}
    \begin{tabularx}{\linewidth}{@{}p{0.20\linewidth}Xp{0.19\linewidth}@{}}
        \toprule
        Outcome & Criterion & Scoring treatment \\
        \midrule
        Success & Native evaluator accepts the required terminal state or target node & Positive outcome \\
        Method failure & A valid run produces a rejected action/state, exhausts its protocol, or terminates without an admissible result & Included as failure \\
        Benchmark failure & Released asset, schema, reset, oracle, or evaluator contract is inconsistent & Quarantined \\
        Infrastructure failure & Required service, container, endpoint, or evaluator process is unavailable independently of system behavior & Excluded and rerun \\
        \bottomrule
    \end{tabularx}
\end{table}

\end{document}